# COLOR IMAGE ENHANCEMENT IN THE FRAMEWORK OF LOGARITHMIC MODELS

Vasile PĂTRAȘCU, Vasile BUZULOIU
Image Processing and Analysis Laboratory (LAPI),
Faculty of Electronics and Telecommunications,
" Politehnica" University Bucharest
E-mail: vpatrascu@tarom.ro, buzuloiu@alpha.imag.pub.ro

**ABSTRACT**
In this paper, we propose a mathematical model for color image processing. It is a logarithmical one. We consider the cube $(-1,1)^3$ as the set of values for the color space. We define two operations: addition $\langle + \rangle$ and real scalar multiplication $\langle \times \rangle$. With these operations the space of colors becomes a real vector space. Then, defining the scalar product $(.|.)$ and the norm $\| \cdot \|$, we obtain a (logarithmic) Euclidean space. We show how we can use this model for color image enhancement and we present some experimental results.

## I. INTRODUCTION

Image enhancement is an important branch of image processing. Its definition according to [5,10], can be expressed as follows: It consists of a collection of techniques that seek to improve the visual appearance of an image, or to convert the image to a form better suited for analysis by a human or a machine. This area of image processing has been and is always an important field of the research in which publications of new techniques and methods take a large place in the literature. Many approaches exist (contrast manipulation, histogram modification, filtering,..) that are well exposed in reference books such as [1-3, 10, 11]. A great number of enhancement methods exist because each application is specific and needs an adapted method [2]. So, the physical nature of images to be processed is of central importance and the need of an adequate image mathematical model appears clearly as a necessity. Stockham [12] proposed an image enhancement method based on the homomorphic theory introduced by Oppenheim [6] and applied to images obtained by transmitted or reflected light. The key of this approach is that it uses an adequate mathematical homomorphism, that performs a transformation in order to use the classical linear image processing techniques. Another approach exists in the general setting of logarithmic representations suited for the imaging processes obtained by transmitted light or for the human visual perception. Jourlin and Pinoli introduced a mathematical framework for this kind of "non-linear" representation [4, 5, 9].

The both models work with semi-bounded sets of values. This is the principal reason that limits their extension to color image processing. The logarithmic model for gray level images presented in [7, 8] does work with bounded real sets. The extension of our model to color images is natural and we develop it here. We consider the unit cube of colors $(0,1)^3$ and we transform it in the cube $(-1,1)^3$ by the simplest linear transformation which translates the point $(0.5, 0.5, 0.5)$ to $(0, 0, 0)$. It is this new cube which plays the central role in our model: it is the space of colors which will be endowed with an Euclidean structure. The remainder of the paper is organized as follows: Section 2 introduces the addition, the real scalar multiplication, the scalar product and the norm for the color space. Similarly, the Section 3 introduces the addition, the real scalar multiplication, the scalar product and the norm for the color image space. Section 4 presents two optimal image transforms using our mathematic model. Section 5 presents experimental results and Section 6 outlines the conclusions.

## II. THE REAL VECTOR SPACE OF THE COLORS

We consider as the space of colors, the cube $E_3=(-1,1)\times(-1,1)\times(-1,1)$. The three components of a vector $v \in E_3$, we will note with r, g and b (red, green, blue). In the cube $E_3$ we will define the addition $\langle + \rangle$ and the real scalar multiplication $\langle \times \rangle$.

*A. Addition*

$\forall v_1, v_2 \in E_3$ with $v_1=(r_1,g_1,b_1)$, $v_2=(r_2,g_2,b_2)$, the sum $v_1 \langle + \rangle v_2$ is defined by the following relation:

$$v_1 \langle + \rangle v_2 = \left( \frac{r_1+r_2}{1+r_1 \cdot r_2}, \frac{g_1+g_2}{1+g_1 \cdot g_2}, \frac{b_1+b_2}{1+b_1 \cdot b_2} \right) \quad (1)$$

The neutral element for addition is $\theta=(0,0,0)$. Each element $v=(r,g,b) \in E_3$ has as its opposite the element $w=(-r,-g,-b)$ and w verifies the following equation: $v \langle + \rangle w = \theta$.

The addition $\langle + \rangle$ is stable, associative, commutative, has a neutral element and each element has an opposite. It results that this operation establishes on $E_3$ a commutative group structure. We can also define the subtraction operation $\langle - \rangle$ by:

$$v_1 \langle - \rangle v_2 = \left( \frac{r_1-r_2}{1-r_1 \cdot r_2}, \frac{g_1-g_2}{1-g_1 \cdot g_2}, \frac{b_1-b_2}{1-b_1 \cdot b_2} \right) \quad (2)$$

With subtraction $\langle - \rangle$, we will note by $\langle - \rangle v$ the opposite of v ( w above ).

*B. Scalar multiplication*

$\forall \lambda \in R, \forall v=(r,g,b) \in E_3$, we define the product between $\lambda$ and v by: $\lambda \langle \times \rangle v =$

$$= \left( \frac{(1+r)^\lambda - (1-r)^\lambda}{(1+r)^\lambda + (1-r)^\lambda}, \frac{(1+g)^\lambda - (1-g)^\lambda}{(1+g)^\lambda + (1-g)^\lambda}, \frac{(1+b)^\lambda - (1-b)^\lambda}{(1+b)^\lambda + (1-b)^\lambda} \right)$$

(3)

The two operations: addition $\langle+\rangle$ and scalar multiplication $\langle\times\rangle$ establish on $E_3$ a real vector space structure.

*C. The Euclidean space of the colors*

We define the scalar product $(.|.)_{E_3}: E_3 \times E_3 \to R$ by:
$\forall v_1, v_2 \in E_3$ with $v_1 = (r_1, g_1, b_1)$, $v_2 = (r_2, g_2, b_2)$,
$$(v_1 | v_2)_{E_3} = \varphi(r_1) \cdot \varphi(r_2) + \varphi(g_1) \cdot \varphi(g_2) + \varphi(b_1) \cdot \varphi(b_2) \quad (4)$$

where $\varphi: (-1,1) \to R$ and $\varphi(x) = \text{arcth}(x)$.

The norm is defined in the normal way:
$\|.\|_{E_3}: E_3 \to [0,\infty)$, $\forall v = (r,g,b) \in E_3$,
$$\|v\|_{E_3} = \sqrt{\varphi^2(r) + \varphi^2(g) + \varphi^2(b)} \quad (5)$$

Our color space becomes a three-dimensional Euclidean space.

### III. THE VECTOR SPACE OF THE COLOR IMAGES

A color image is a function defined on a bi-dimensional compact D from $R^2$ taking the values in the color space $E_3$. We note with $F(D,E_3)$ the set of color images defined on D. We can extend the operations and the functions from color space $E_3$ to color images $F(D,E_3)$ in a natural way.

*A. Addition*

$\forall f_1, f_2 \in F(D,E_3)$, $\forall (x,y) \in D$,
$$(f_1 \langle+\rangle f_2)(x,y) = f_1(x,y) \langle+\rangle f_2(x,y) \quad (6)$$

The neutral element is the identical null function. The addition $\langle+\rangle$ is stable, associative, commutative, has a neutral element and each element has an opposite. As a conclusion, this operation establishes on the set $F(D,E_3)$ a commutative group structure.

*B. Scalar multiplication*

$\forall \lambda \in R, \forall f \in F(D,E_3)$, $\forall (x,y) \in D$,
$$(\lambda \langle\times\rangle f)(x,y) = \lambda \langle\times\rangle f(x,y) \quad (7)$$

The two operations, addition $\langle+\rangle$ and scalar multiplication $\langle\times\rangle$ establish on $F(D,E_3)$ a real vector space structure.

*C. The Hilbert space of the color images*

Let be f and g two integrable functions from $F(D,E_3)$. We can define the scalar product by: $\forall f_1, f_2 \in F(D,E_3)$
$$(f_1 | f_2)_{L^2(E_3)} = \int_D (f_1(x,y) | f_2(x,y))_{E_3} dxdy \quad (8)$$

Further on, we will define the norm.

$$\forall f \in F(D,E_3), \quad \|f\|_{L^2(E_3)} = \left(\int_D \|f(x,y)\|^2_{E_3} dxdy\right)^{\frac{1}{2}} \quad (9)$$

Thus, the color images space $F(D,E_3)$ becomes a Hilbert space.

### IV. APPLICATIONS IN ENHANCEMENT OF COLOR IMAGES

The enhancement algorithms that we will present in this paper consists of affine transformations, $t: F(D,E_3) \to F(D,E_3)$ given by the equation:

$$t(f) = \alpha \langle\times\rangle f \langle+\rangle \beta \langle\times\rangle k \quad (10)$$

where f is the original image and $h = t(f)$ is the enhanced image. Here k is a constant image. The parameters $\alpha, \beta$ will be determined so that the transformation satisfies some conditions. We shall study two sets of conditions. Firstly, we make some preliminary notations. Let be $f: D \to E_3$ a color image, having the scalar components $f_r: D \to E$, $f_g: D \to E$, $f_b: D \to E$.

We consider the following subset of the support D:
$D_{r1} = \{(x,y) \in D | f_r(x,y) \leq m(f_r(D))\}$
$D_{r2} = \{(x,y) \in D | f_r(x,y) \geq m(f_r(D))\}$
$D_{g1} = \{(x,y) \in D | f_g(x,y) \leq m(f_g(D))\}$
$D_{g2} = \{(x,y) \in D | f_g(x,y) \geq m(f_g(D))\}$
$D_{b1} = \{(x,y) \in D | f_b(x,y) \leq m(f_b(D))\}$
$D_{b2} = \{(x,y) \in D | f_b(x,y) \geq m(f_b(D))\}$

where $m(f_i(D)) = \dfrac{1}{\text{area}(D)} \int_D f_i(x,y) dxdy$.

In the discrete case, area(D) will be replaced by card(D). We will consider the following vectors:
$v_0 = (m(f_r(D)), m(f_g(D)), m(f_b(D)))$
$v_1 = (m(f_r(D_{r1})), m(f_g(D_{g1})), m(f_b(D_{b1})))$
$v_2 = (m(f_r(D_{r2})), m(f_g(D_{g2})), m(f_b(D_{b2})))$
$u_L = (\text{th}(1), \text{th}(1), \text{th}(1))$
$w_0 = (0, 0, 0)$, $w_1 = (-0.5, -0.5, -0.5)$,
$w_2 = (0.5, 0.5, 0.5)$

*The algorithm A:* a constant translation on the each component of the color.
In this case we put:
$$\forall x, y \in D, \quad k(x,y) = u_L \quad (11)$$

We obtain the parameters $\alpha, \beta$ from the over-determined linear system:
$$\begin{cases} \alpha \langle\times\rangle v_0 \langle+\rangle \beta \langle\times\rangle u_L = w_0 \\ \alpha \langle\times\rangle (v_1 \langle-\rangle v_0) = w_1 \\ \alpha \langle\times\rangle (v_2 \langle-\rangle v_0) = w_2 \end{cases} \quad (12)$$

Let us use the notations:
$c_{vv} = \|v_0\|^2_{E_3} + \|v_1 \langle-\rangle v_0\|^2_{E_3} + \|v_2 \langle-\rangle v_0\|^2_{E_3}$,
$c_{uu} = \|u_L\|^2_{E_3}$, $c_{vu} = (v_0 | u_L)_{E_3}$, $c_{uw} = (u_L | w_0)_{E_3}$
$c_{vw} = (v_0 | w_0)_{E_3} + (v_1 | w_1)_{E_3} + (v_2 | w_2)_{E_3}$

Then the (generalized) solution (i.e. corresponding to the minimum mean square error) is:
$$\alpha = \frac{c_{vw} \cdot c_{uu}}{c_{vv} \cdot c_{uu} - (c_{vu})^2} \quad (13)$$

$$\beta = \frac{-c_{vw} \cdot c_{vu}}{c_{vv} \cdot c_{uu} - (c_{vu})^2} \quad (14)$$

and the transformation is:

$$t(f) = \frac{c_{vw} \cdot c_{uu}}{c_{vv} \cdot c_{uu} - (c_{vu})^2} \langle \times \rangle \left( f \langle - \rangle \frac{c_{vu}}{c_{uu}} \langle \times \rangle u_L \right) \quad (15)$$

*The algorithm B:* a translation proportional to the mean. We suppose $\|v_0\|_{E_3} > 0$. In this case we shall choose the constant vector $k = v_0$ (16)
The over-determined system from which the parameters $\alpha, \beta$ are to be found is:

$$\begin{cases} \alpha \langle \times \rangle v_0 \langle + \rangle \beta \langle \times \rangle v_0 = w_0 \\ \alpha \langle \times \rangle v_1 \langle + \rangle \beta \langle \times \rangle v_0 = w_1 \\ \alpha \langle \times \rangle v_2 \langle + \rangle \beta \langle \times \rangle v_0 = w_2 \end{cases} \quad (17)$$

We again use some notations:
$c_{vv} = \|v_0\|^2_{E_3} + \|v_1\|^2_{E_3} + \|v_2\|^2_{E_3}$, $c_{uu} = 3 \cdot \|v_0\|^2_{E_3}$
$c_{vu} = (v_0|v_0)_{E_3} + (v_1|v_0)_{E_3} + (v_2|v_0)_{E_3}$
$c_{vw} = (v_0|w_0)_{E_3} + (v_1|w_1)_{E_3} + (v_2|w_2)_{E_3}$
$c_{uw} = (v_0|w_0)_{E_3} + (v_0|w_1)_{E_3} + (v_0|w_2)_{E_3}$
and the minimum mean square error solution is:

$$\alpha = \frac{c_{vw} \cdot c_{uu}}{c_{vv} \cdot c_{uu} - (c_{vu})^2} \quad (18)$$

$$\beta = \frac{-c_{vw} \cdot c_{vu}}{c_{vv} \cdot c_{uu} - (c_{vu})^2} \quad (19)$$

and the transformation is:

$$t(f) = \frac{c_{vw} \cdot c_{uu}}{c_{vv} \cdot c_{uu} - (c_{vu})^2} \langle \times \rangle \left( f \langle - \rangle \frac{c_{vu}}{c_{uu}} \langle \times \rangle v_0 \right) \quad (20)$$

## V. EXPERIMENTAL RESULTS

We exemplify our algorithms on more or less standard test images: "couple" (Fig.1), "fruit" (Fig.4), "kidsat3" (Fig. 7) and "boat" (Fig. 10).
For the "couple" image we get the following means:
$v_0 = (-0.705, -0.784, -0.784)$
$v_1 = (-0.868, -0.883, -0.873)$
$v_2 = (-0.471, -0.567, -0.612)$
For the algorithm A one obtains:
$\alpha = 1.434, \beta = 1.429, k = (0.762, 0.762, 0.762)$
$t(f) = 1.434 \langle \times \rangle f \langle + \rangle 1.429 \langle \times \rangle k$

For the algorithm B the values are:
$\alpha = 1.472, \beta = -1.461, k = (-0.705, -0.784, -0.784)$
$t(f) = 1.472 \langle \times \rangle f \langle - \rangle 1.461 \langle \times \rangle k$

Enhanced images for "couple" are showed in Fig.2 for algorithm A and in Fig.3 for algorithm B.

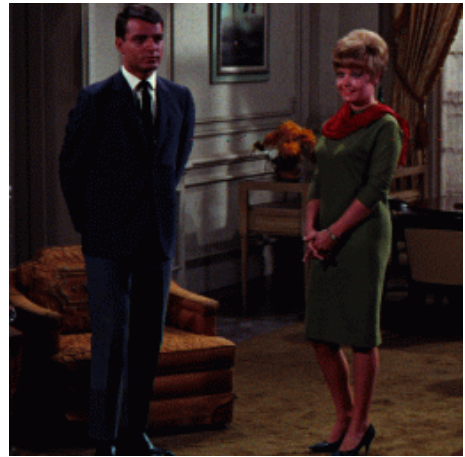

Fig.1 Original image "couple"

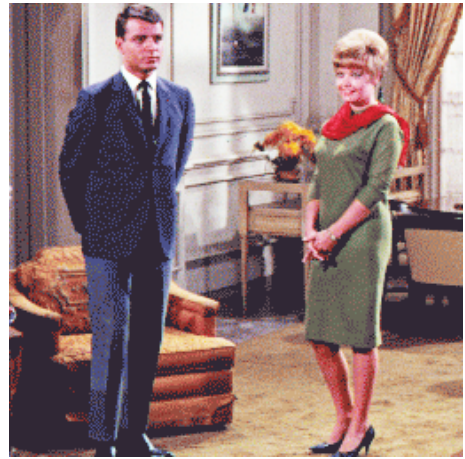

Fig.2 Enhanced image ("couple") with algorithm A

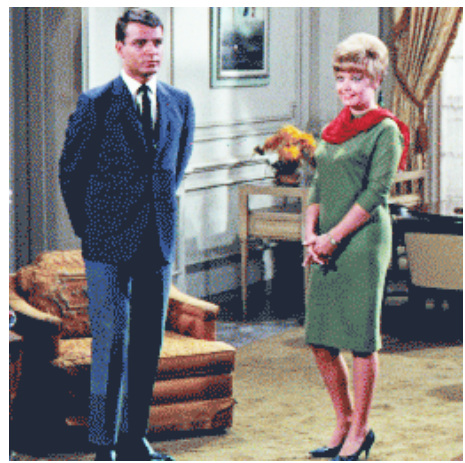

Fig.3 Enhanced image ("couple") with algorithm B

For the "fruit" image we get the following values:
$$v_0=(-0.187, -0.388, -0.586)$$
$$v_1=(-0.549, -0.710, -0.785)$$
$$v_2=( 0.359,  0.079, -0.263)$$

The algorithm A gives:
$\alpha = 1.081$, $\beta = 0.458$, k=(0.762, 0.762, 0.762)
$$t(f) = 1.081 \langle \times \rangle f \langle + \rangle 0.458 \langle \times \rangle k$$

The algorithm B gives:
$\alpha = 1.181$, $\beta = -1.156$, k=(-0.187, -0.388, -0.586)
$$t(f) = 1.181 \langle \times \rangle f \langle - \rangle 1.156 \langle \times \rangle k$$

Enhanced images for "fruit" are showed in Fig.5 for algorithm A and in Fig.6 for algorithm B.

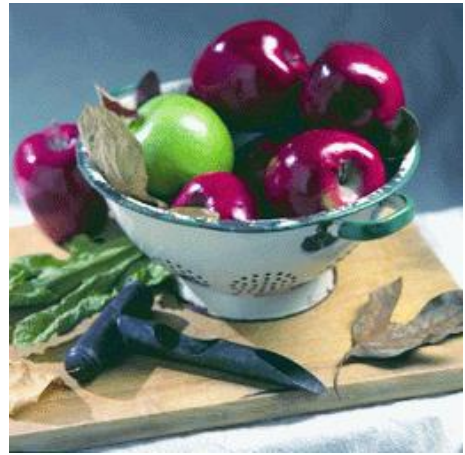

Fig.6 Enhanced image ("fruit") with algorithm B

For the "kidsat3" image we get the following values:
$$v_0=(-0.694, -0.727, -0.580)$$
$$v_1=(-0.868, -0.846, -0.714)$$
$$v_2=(-0.347, -0.515, -0.459)$$

The algorithm A gives:
$\alpha = 1.384$, $\beta = 1.116$, k=(0.762, 0.762, 0.762)
$$t(f) = 1.384 \langle \times \rangle f \langle + \rangle 1.116 \langle \times \rangle k$$

The algorithm B gives:
$\alpha = 1.450$, $\beta = -1.447$, k=(-0.694, -0.727, -0.580)
$$t(f) = 1.450 \langle \times \rangle f \langle - \rangle 1.447 \langle \times \rangle k$$

Enhanced images for "kidsat3" are showed in Fig.8 for algorithm A and in Fig.9 for algorithm B.

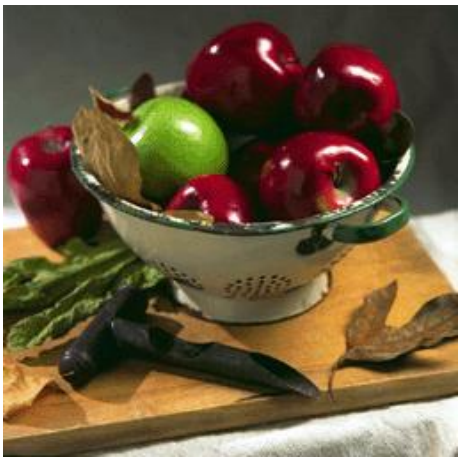

Fig.4 Original image "fruit"

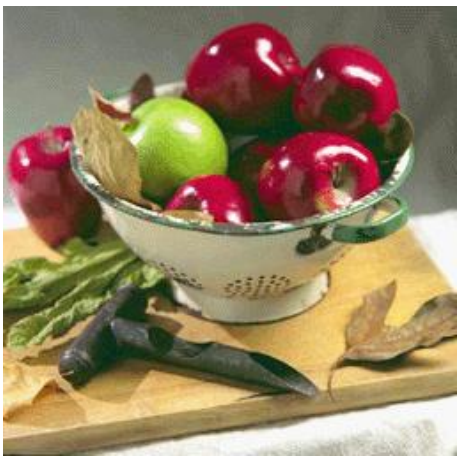

Fig.5 Enhanced image ("fruit") with algorithm A

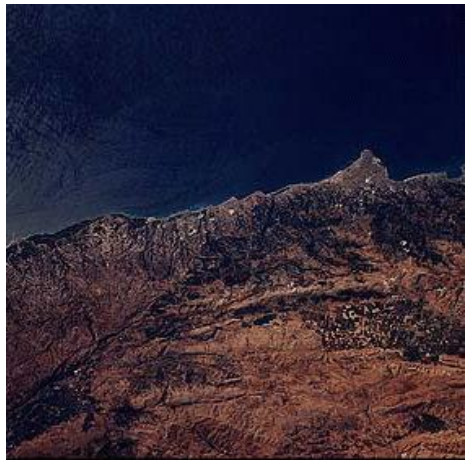

Fig.7 Original image "kidsat3"

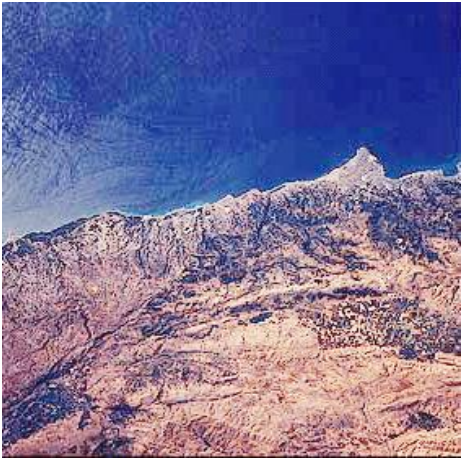

Fig.8 Enhanced image ("kidsat3") with algorithm A

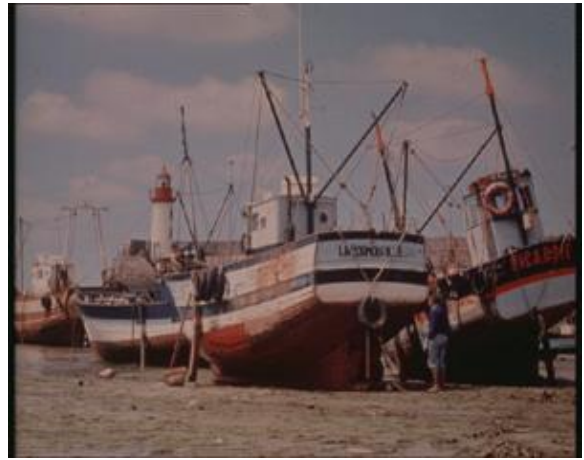

Fig.10 Original image "boat"

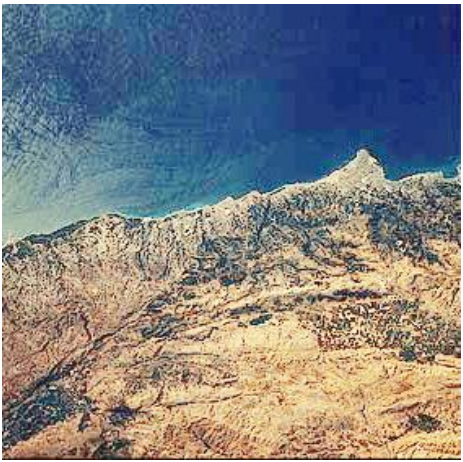

Fig.9 Enhanced image ("kidsat3") with algorithm B

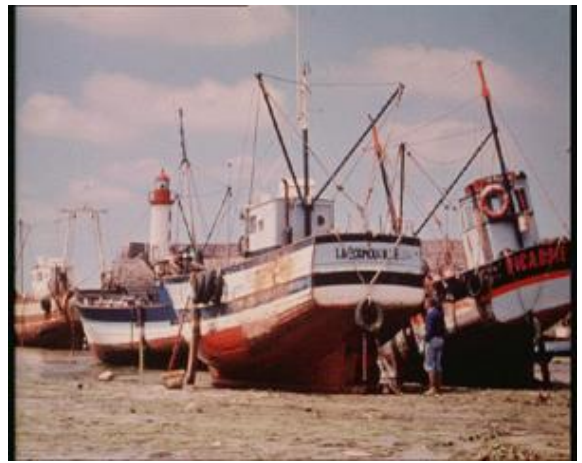

Fig.11 Enhanced image ("boat") with algorithm A

For the "boat" image we obtained the following values:
$$v_0 = (-0.194, -0.323, -0.338)$$
$$v_1 = (-0.557, -0.675, -0.676)$$
$$v_2 = (-0.001, -0.119, -0.107)$$

The algorithm A gives:
$\alpha = 1.402, \beta = 0.412, k = (0.762, 0.762, 0.762)$
$$t(f) = 1.402 \langle \times \rangle f \langle + \rangle 0.412 \langle \times \rangle k$$

The algorithm B gives:
$\alpha = 1.538, \beta = -1.941, k = (-0.194, -0.323, -0.338)$
$$t(f) = 1.538 \langle \times \rangle f \langle - \rangle 1.941 \langle \times \rangle k$$

Enhanced images for "boat" are showed in Fig.11 for algorithm A and in Fig.12 for algorithm B.

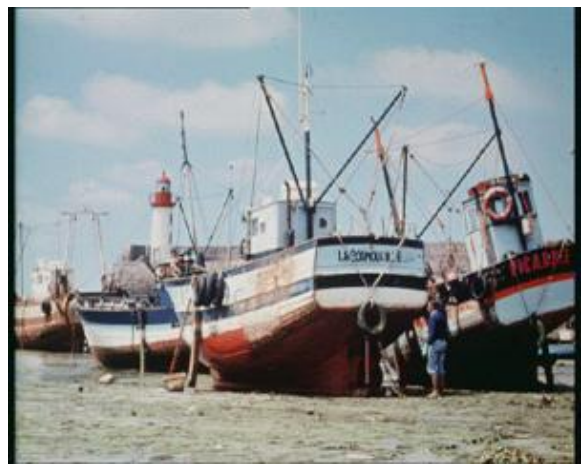

Fig.12 Enhanced image ("boat") with algorithm B

## VI. CONCLUSIONS

Using this mathematical model we obtained two very simple algorithms for color image enhancement. The algorithm A - formulae (11, 13, 14, and 15) - improves the color contrast and saturation and preserves quite well the hue, while the algorithm B - formulae (16, 18, 19, and 20) – tends to attenuate the dominant colors which results in slight hue modifications. Nevertheless these modifications in some cases produce even better bright images (see Fig. 12). The examples above prove that the mathematical framework of the Euclidean color space that we built, is very useful for image processing: it allows one to quickly define a large set of algorithms.

The main advantage consists in the ability to parameterize the algorithms and to simply introduce optimality criteria. On the other hand, the computing involved is kept at the lowest level (a linear transformation for each pixel).